\documentclass[aip,jcp,amsmath,amssymb,floatfix,reprint,citeautoscript,noeprint,longbibliography,nofootinbib]{revtex4-2}

\usepackage[utf8]{inputenc}
\usepackage{graphicx}
\usepackage{wrapfig}
\usepackage{placeins}
\usepackage{bibunits}
\usepackage{siunitx}
\usepackage[colorlinks,allcolors=blue,citecolor=blue,urlcolor=blue]{hyperref}
\usepackage[mathlines]{lineno}

\makeatletter
\renewcommand{\fnum@figure}{\textbf{Figure \thefigure}}
\makeatother

\usepackage{etoolbox}
\makeatletter
\patchcmd{\@makecaption}{\centering}{\raggedright}{}{}
\renewcommand{\fnum@figure}{\sffamily\figurename~\thefigure}
\renewcommand{\fnum@table}{\sffamily\tablename~\thetable}
\patchcmd{\@makecaption}{\small}{\small\sffamily}{}{}
\makeatother

\graphicspath{{figures/}}
\hypersetup{colorlinks=true, urlcolor=blue}

\defaultbibliography{references}
\defaultbibliographystyle{aipnum4-2}

\begin{document}

\def\mytitle{Cutting Through the Noise: On-the-fly Outlier Detection for Robust Training of Machine Learning Interatomic Potentials}
\title{\mytitle}

\author{Terry C. W. Lam}
\affiliation{Cavendish Laboratory, Department of Physics, University of Cambridge, Cambridge, CB3 0HE, United Kingdom}
\affiliation{Lennard-Jones Centre, University of Cambridge, Trinity Lane, Cambridge, CB2 1TN, United Kingdom}

\author{Niamh O’Neill}
\affiliation{Yusuf Hamied Department of Chemistry, University of Cambridge, Lensfield Road, Cambridge, CB2 1EW, United Kingdom}
\affiliation{Cavendish Laboratory, Department of Physics, University of Cambridge, Cambridge, CB3 0HE, United Kingdom}
\affiliation{Lennard-Jones Centre, University of Cambridge, Trinity Lane, Cambridge, CB2 1TN, United Kingdom}

\author{Christoph Schran}
\email{cs2121@cam.ac.uk}
\affiliation{Cavendish Laboratory, Department of Physics, University of Cambridge, Cambridge, CB3 0HE, United Kingdom}
\affiliation{Lennard-Jones Centre, University of Cambridge, Trinity Lane, Cambridge, CB2 1TN, United Kingdom}

\author{Lars L. Schaaf}
\email{lls34@cam.ac.uk}
\affiliation{Cavendish Laboratory, Department of Physics, University of Cambridge, Cambridge, CB3 0HE, United Kingdom}
\affiliation{Lennard-Jones Centre, University of Cambridge, Trinity Lane, Cambridge, CB2 1TN, United Kingdom}

\date{\today}

\begin{abstract}
The accuracy of machine learning interatomic potentials suffers from reference data that contains numerical noise. Often originating from unconverged or inconsistent electronic-structure calculations, this noise is challenging to identify.
Existing mitigation strategies such as manual filtering or iterative refinement of outliers,  require either substantial expert effort or multiple expensive retraining cycles, making them difficult to scale to large datasets.
Here, we introduce an on-the-fly outlier detection scheme that automatically down-weights noisy samples, without requiring additional reference calculations.
By tracking the loss distribution via an exponential moving average, this unsupervised method identifies outliers throughout a single training run.
We show that this approach prevents overfitting and matches the performance of iterative refinement baselines with significantly reduced overhead.
The method's effectiveness is demonstrated by recovering accurate physical observables for liquid water from unconverged reference data, including diffusion coefficients.
Furthermore, we validate its scalability by training a foundation model for organic chemistry on the SPICE dataset, where it reduces energy errors by a factor of three.
This framework provides a simple, automated solution for training robust models on imperfect datasets across dataset sizes.
\end{abstract}

{\maketitle}

\begin{bibunit}

\section{Introduction}

Machine learning interatomic potentials (MLIPs) are routinely employed to circumvent the high computational cost of \textit{ab initio} quantum chemistry methods, with minimal sacrifice in accuracy \cite{MLIP1, MLIP2, MLIP3, MLIP4, MLIP5, MLIP6, MLIP7, MLIP8, Thiemann_2025}.
Trained on reference data, MLIPs predict the potential energy of an atomic configuration directly from
the constituent atoms' positions and elemental identities.
By bypassing the need to solve the electronic Schrödinger equation (or approximations thereof) at each time step, MLIPs can accelerate atomic simulations by orders of magnitude while generally exhibiting linear scaling with system size~\cite{Bart_k_2017, Unke_2021, Thiemann_2025}. 

Previously, significant effort was required to curate relevant training datasets.
To automate the procedure, active learning approaches were often employed to iteratively improve models~\cite{Smith2018, Podryabinkin2019, DPGEN, Schran2020, Vandermause2020, Sivaraman2020, Schran2021, Young2021, Vandermause2022, Schaaf2023, vanderOord2023}.
This process typically involves computing reference energies and forces for new configurations sampled from simulations run with preliminary versions of the MLIP.
This approach generally resulted in models trained for specific applications~\cite{Deringer2019, Novikov2021, Schaaf2023, Rhodes2024, Behler2025}. 

Recently, a new paradigm has emerged with models trained on large, diverse reference datasets, resulting in so-called `foundation models'\cite{CHGNet, M3GNet, GNoME, GRACE, EquiformerV2, MACEMP0, MACEOFF, orbnet, uma, Ple2025, batatia2025cross}.
These models offer strong out-of-the-box performance across a wide range of applications, from materials science to organic chemistry.
Furthermore, they have demonstrated strong generalisability.
For instance, models trained exclusively on small crystal structures can successfully simulate systems with structural and chemical complexity far exceeding the training data, such as carbon capture in porous materials or peptides in aqueous solution~\cite{MACEMP0}.
For additional accuracy, foundation models can be fine-tuned on a small set of configurations, requiring less data than training a model from scratch~\cite{Batatia2025}.

Despite this progress, MLIPs often struggle when trained on noisy data, which is becoming even more problematic in the large data regime~\cite{SPICE1, SPICE2, Omol25, revMD17, OC20}.
Such noise can be random or systematic, frequently originating from numerical errors in the reference electronic structure method or mismatches in the electronic structure settings for different systems.
The sources of these errors are manifold and method-dependent.
For example, reference calculations based on density functional theory (DFT) may suffer from incomplete self-consistent field (SCF) convergence~\cite{Lejaeghere_2016, Kuryla_2025}.
Concurrently, other high-accuracy methods, such as variational (VMC) and diffusion (DMC) Monte Carlo, are intrinsically stochastic and introduce inherent noise into the reference forces and energies~\cite{Andrea_2015, Slootman2024}.
In the context of foundation models, it is difficult to avoid all sources of noise when the dataset sizes exceeds millions of configurations. 

Current strategies to mitigate the impact of noisy data primarily involve manual filtering \cite{Kuryla_2025, Omol25, MACEOFF} or iterative refinement \cite{MACEMP0, MACEOFF}.
Manual filtering is labour-intensive, requires domain-specific chemical expertise, and risks the erroneous removal of valuable data points~\cite{MACEOFF}.
In the context of foundation models trained on hundreds of millions of configurations, such a manual approach quickly becomes infeasible.
An alternative strategy involves training an initial MLIP on the entire dataset and subsequently filtering configurations on which the model exhibits high prediction errors.
The underlying assumption is that the MLIP, constrained by smoothness, will fit the average data behavior and thus identify noisy configurations as high-error outliers.
This method functions for us as an iterative refinement baseline\cite{MACEMP0, MACEOFF}.
Iterative refinement, however, is computationally expensive as it requires repeated training cycles on the same large datasets.

\begin{figure}[tb]
    \centering
    \includegraphics[width=\linewidth]{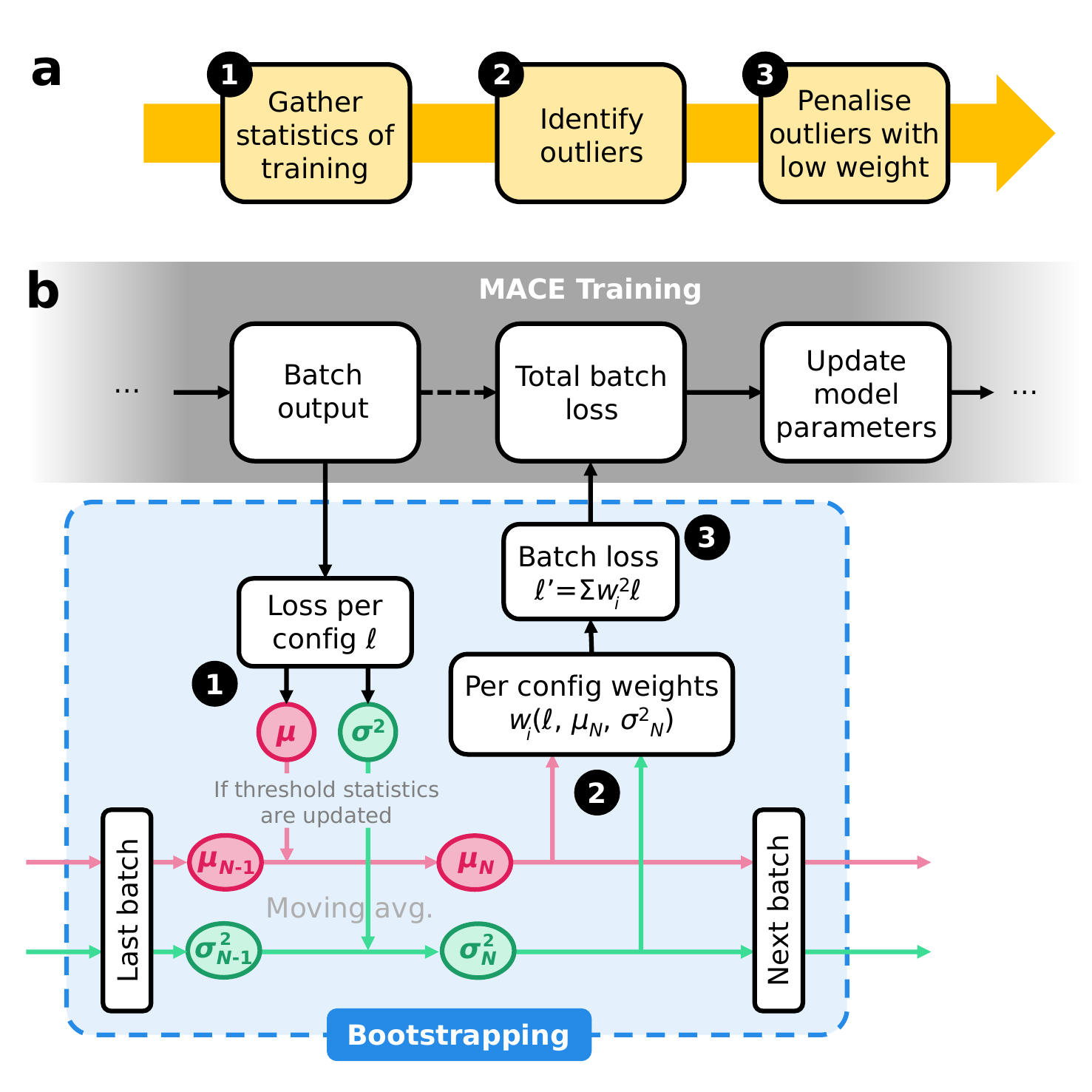}
    \vspace{-.8cm}
    \caption{%
    \textbf{On-the-fly outlier detection.} (a) A high level overview of the steps for detecting outliers in the training data and down weighting their impact on the training loss. (b) A overview of the changes in the training procedure, showing normal MLIP training steps (top) and the new on-the-fly bootstrapping technique (bottom). Number labels indicate which parts of the training modifications correspond to the high level overview in (a).}
    \label{fig:schematic}
\end{figure}

In this paper, we introduce an on-the-fly outlier detection approach that requires neither model refitting, nor chemical expertise.
The core concept is to systematically down-weight suspected noisy data points within each batch during training.
A data point's probability of being an outlier is evaluated based on its deviation from a tracked error distribution, which is maintained using an exponential moving average (EMA) of batch errors (e.g., being $3\sigma$ outside the EMA-derived mean).
This strategy enables noise-resilient fitting in a single training run.
For our development, we draw inspiration from applications in the broader machine learning community~\cite{dynamic_bootstrap, Bootstrap_2024, Reed_2015, Han2018, Song_2023, bukharin_2023}.
In this context, our approach can be classified as a dynamic bootstrapping technique \cite{dynamic_bootstrap}, where loss weights vary between one (full trust in the reference data) and zero, depending on the assessed likelihood of the label being noisy.

We find that this noise detection method reduces overfitting, leads to improved physical observables, and accelerates training in the foundation model regime.
We test the approach on marginally noisy data, finding that standard `vanilla' training leads to a higher validation loss than our proposed method.
When the training set is re-evaluated with more converged reference DFT settings, our model, trained on the original noisy data, accurately predicts energies and forces when compared to the converged data.
Furthermore, we show that the improvements in validation root-mean-square error (RMSE) directly impacts the accuracy of observables derived from simulations, such as calculated diffusion coefficients.
Finally, we apply the approach to training an organic force field foundation model on the SPICE 2.0 dataset \cite{SPICE1, SPICE2}, where the resulting model achieves energy errors a factor of three lower than the vanilla model.
This automated noise-resilience technique provides a practical solution for both training large-scale, next-generation foundation models from scratch and for performing robust fine-tuning, thereby accelerating Machine-learning(ML)-driven molecular and materials discovery.

\section{Method}
\subsection{Overview of Noise-resilient Training}

The proposed on-the-fly outlier detection for training noise-resilient MLIPs consists of three steps, as visualised in Fig.~\ref{fig:schematic}.
Firstly, we track the average and standard deviation of the loss throughout training.
Metrics for the loss distribution as a function of epoch are needed for determining outliers.
Secondly, at each batch we determine the likelihood of a given configuration being an outlier based on the training example's loss compared to the determined loss distribution of previous examples.
The higher the loss compared to the mean and standard deviation, the more likely a configuration is assumed to have been mislabelled. 
Finally, we reweigh all training configurations based on their probability of being noisy data, before computing the loss.
This way the impact of noisy outliers on training performance is mitigated
Below we outline each of these steps in detail. 

\begin{figure*}[t!]
    \centering
    \includegraphics[width=.9\linewidth]{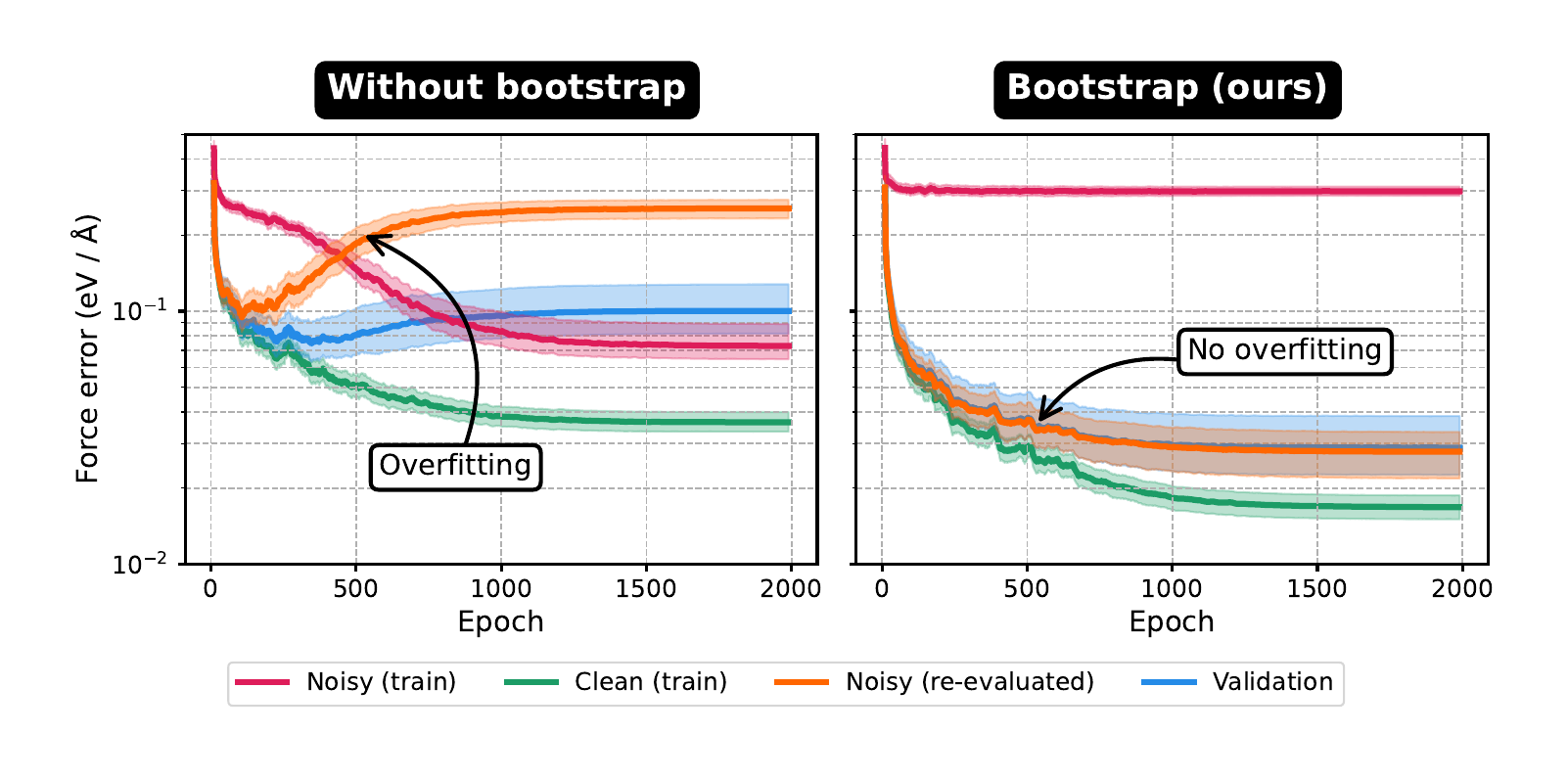}
    \vspace{-.7cm}
    \caption{%
    \textbf{Bootstrapping prevents overfitting.}
    Force error curves of the noisy samples (red), clean samples (green), and samples in the validation set (blue), as a function of epoch, without and with bootstrapping applied.  Thick line represents median and shaded area represents inter-quartile range (IQR).  This training error is defined as the RMSE with respect to the reference (potentially false) forces the training sees. The noisy training data is re-evaluated with more converged reference method. Comparisons between the models predictions and the unseen re-evaluated noisy configurations (orange), indicates that the noise resilient model is able to predict the true forces at validation accuracy.
    }
    \label{fig:overfit}
\end{figure*}

Our implementation of on-the-fly noise detection arises from the key consideration that outliers take longer  to learn \cite{Arpit2017, Zhang_2017, dynamic_bootstrap}.
We show this empirically in Fig.~\ref{fig:overfit}, where the distinct splitting of noisy and clean training data shows key differences in training dynamics.
We explore this behaviour in further detail in the results section.

\begin{figure}[tb!]
    \centering
    \includegraphics[width=\linewidth]{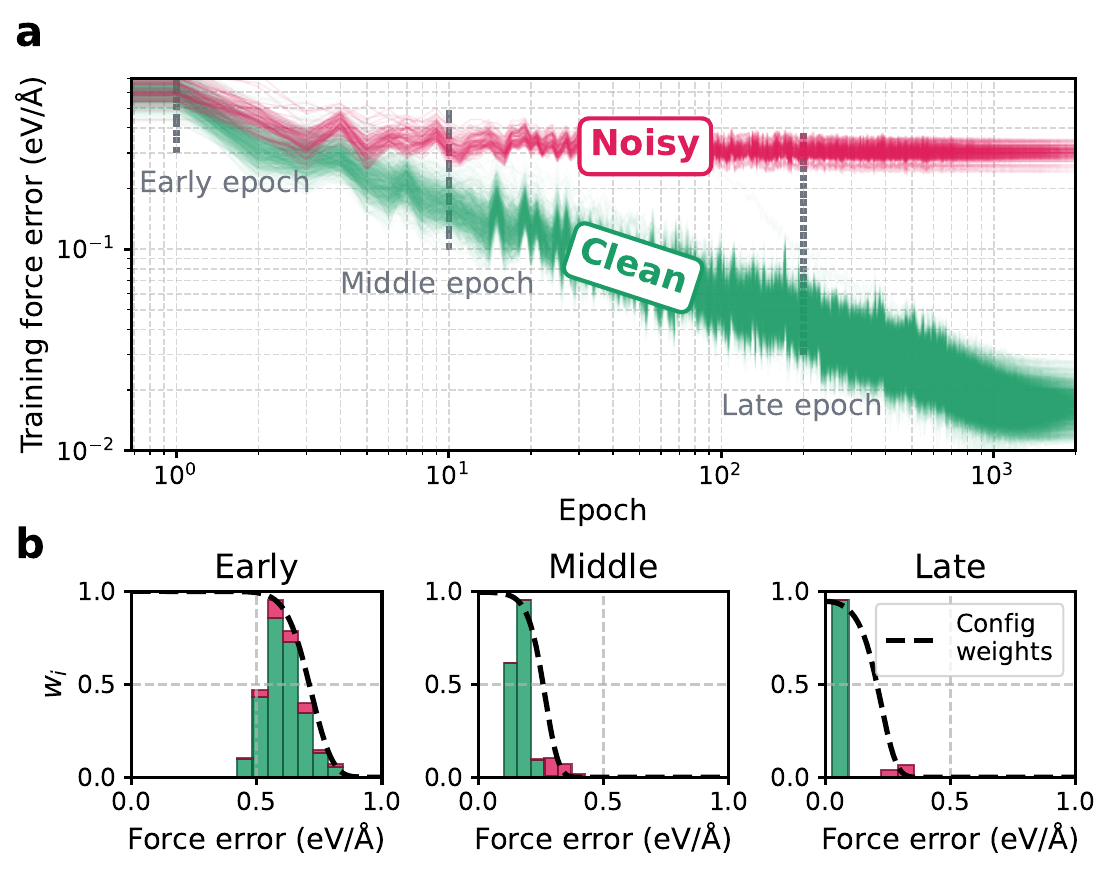}\\
    \caption{%
    \textbf{Noise-resilient training on a revMD17 dataset.}
    (a) Log-log plots of the error evolution with epoch in the 2000-epoch training.  Each line represents one configuration, and are coloured according to whether the labels were from revMD17 (`clean', green), or from MD17 (`noisy', red). (b) Force error distribution at early (${\sim}10^0$), middle (${\sim}10^1$) and late epochs (${\sim}10^2$).  The assignment of the bootstrapping weights based on the force error shown.}
    \label{fig:weights}
\end{figure}

\subsection{Training Loss Modification}

The standard training objective for an MLIP minimizes a loss function $\mathcal{L}_{\text{total}}$, which is a weighted sum of various observables, usually including energy ($\ell_E$), force ($\ell_F$), and stress ($\ell_S$) components.
For a batch of $N_B$ configurations, this is expressed as:
\begin{align}
    \mathcal{L}_{\text{total}} &= \frac{1}{N_B} \sum_{i=1}^{N_B} \mathcal{L}_i \\ 
    &= \frac{1}{N_B} \sum_{i=1}^{N_B} \left( \lambda_E \ell_{E,i} + \lambda_F \ell_{F,i} + \lambda_S \ell_{S,i} \right)
    \label{eq:standard_loss}
\end{align}
where $\mathcal{L}_i$ is the total loss for configuration $i$, and $\lambda$ terms are hyper-parameters that balance the contribution of each component (e.g., $\ell_{F,i}$ is the mean squared error of the forces for configuration $i$).

To mitigate overfitting on noisy reference data, we introduce a dynamic bootstrapping scheme \cite{dynamic_bootstrap, Reed_2015}.
This modifies the loss function by applying a per-configuration weight, $w_i$, which represents the model's assessed probability that configuration $i$ is ``clean'' (i.e., not an outlier) resulting in the modified loss function, $\mathcal{L}'$,
\begin{align}
    \mathcal{L}' &= \frac{1}{N_B} \sum_{i=1}^{N_B} w_i^2 \mathcal{L}_i .
    \label{eq:weighted_loss}
\end{align}

The squared-weight formulation highlights parallels to fitting on a ``soft" target label~\cite{Reed_2015, Bootstrap_2024}, when the $L_2$ loss is considered.
Fitting on the loss $\mathcal{L}'$ can be interpreted as fitting on $y'_{i,\text{ref}}$, which interpolates between the potentially noisy dataset reference $y_{i,\text{ref}}$ and the model's own prediction $y_{i,\text{pred}}$, such that
\begin{equation}
    y'_{i,\text{ref}} \equiv w_i y_{i,\text{ref}} + (1-w_i)y_{i,\text{pred}}.
\end{equation}

\subsection{On-the-fly Weight Estimation}

The weights $w_i$ are determined on-the-fly during training based on each configuration's loss $\mathcal{L}_i$ relative to the loss distribution across the training dataset.
We model the distribution of loss after $\beta$ batches $P(\mathcal{L})$ as a single Gaussian,
\begin{equation}
    \mathcal{L}_\beta \sim \mathcal{N}(\mu_\beta, \sigma^2_\beta),
\end{equation}
where $\mu_\beta$ and $\sigma^2_\beta$ are the average and variance of the loss over the entire dataset of the model with weights after $\beta$ batches of training.
This allows for simple identification of outliers based on the standard score ($z_{i,\beta}= (\mathcal{L}_{i,\beta}-\mu_\beta)/\sigma_\beta$), where $\mathcal{L}_{i,\beta}$ is the loss for configuration $i$ in batch $\beta$.

Determining the average and variance of the loss across the entire dataset at each step would result in significant overhead.
Using the instantaneous batch mean, with batch sizes as low as 8, results in noisy estimates.
We therefore dynamically track the parameters of this distribution, by maintaining an exponential moving average (EMA) of the mean ($\mu$) and variance ($\sigma^2$) of the batch losses
\begin{align}
    \mu_\beta & = (1-\alpha) \mu_{\beta-1} + \alpha \mu_{\text{batch}}  \label{eq:EMA1} \\ 
    \sigma^2_\beta & = (1-\alpha) \sigma^2_{\beta-1} + \alpha \sigma^2_{\text{batch}} ,\label{eq:EMA2}
\end{align}
where $\beta$ is the batch index, $\mu_{\text{batch}}$ and $\sigma^2_{\text{batch}}$ are the mean and variance of the current batch, and $\alpha$ is the EMA decay rate.
The rate $\alpha$ is set relative to the number of training batches per epoch, $n$ (e.g., $\alpha \approx 0.99$ for $n > 100$), ensuring a smooth average that is robust to transient batch-to-batch fluctuations.
As visible in Appendix \ref{subsec:app-weight-estimation}, this results in a smoothly decaying $\mu$ and $\sigma$ throughout training.

Using the tracked $\mu_\beta$ and $\sigma_\beta$, we compute a $z$-score for each configuration $i$ in the current batch $\beta$: $z_{i,\beta} = (\mathcal{L}_{i,\beta} - \mu_\beta) / \sigma_\beta$.
The weight $w_{i,\beta}$ is then assigned using a smoothed thresholding function based on the Gaussian cumulative distribution function
\begin{equation}
    w_{i,\beta} (z_{i,\beta}) = \frac{1}{2} \left [1 + \text{erf}\left (\frac{z_t -z_{i,\beta}}{\sqrt{2}} \right ) \right ],
    \label{eq:weight_calc}
\end{equation}
where $\text{erf}(x)$ is the error function and $z_t$ is a hyperparameter defining the $z$-score threshold for outlier detection (e.g., $z_t=3$).
This function smoothly transitions $w_{i,\beta}$ from $1$ (for $z_{i,\beta} \ll z_t$, presumed clean) to $0$ (for $z_{i,\beta} \gg z_t$, presumed noisy).
$z_t$ can be chosen to reflect the percentage of outliers and the relative error distribution.
In practice, we find that the performance is not sensitive to the exact setting of $z_t$ (see Appendix \ref{subsec:app-threshold-sensitivity}).

During training, we monitor the batch-averaged weight $\langle w_i \rangle$.
A warning is issued if $\langle w_i \rangle$ drops below a predefined value, ensuring that the mean gradient magnitude remains sufficient for stable model convergence.
As the separation between clean and noisy sample losses often becomes more distinct in later epochs as seen in Fig.~\ref{fig:weights}, these statistics can be updated less frequently than at the start to reduce computational overhead.

Previous approaches in image classification have used more complicated bimodal distributions, to model the loss distribution across a dataset containing outliers\cite{dynamic_bootstrap, DivideMix}.
We find that a simple $z$-score approach avoids numerical instabilities of fitting more flexible distributions such as Gaussian mixture models \cite{Permuter2006} or Beta mixture models \cite{Ma2011}, since the typically outlier abundance is only a small fraction ($\lesssim 10\%$) of the dataset.
A single Gaussian robustly captures the central tendency of the data, allowing high-loss noisy samples to be identified as statistical outliers in the distribution's tail.

While generally applicable to all MLIPs, we have implemented this approach for on-the-fly outlier detection in the MACE framework~\cite{MACE}.
While further details are provided in Sec.~\ref{sec:comp-det}, it is important to note that this implementation can be easily scaled to multi-GPU inference.

\section{Results}

In this section, we evaluate the proposed noise-resilient training approach across a range of tasks and dataset scales.
We begin with controlled benchmarks in which noisy training labels are paired with independently converged reference labels for validation.
We then compare against a standard iterative refinement baseline.
Finally, we assess whether the approach improves downstream physical observables and remains effective in the foundation-model regime.

\subsection{Bootstrapping reduces over-fitting}

To evaluate the effectiveness of noise-resilient training in preventing overfitting, we constructed a synthetic dataset based on the MD17 and revised MD17 (revMD17) datasets.
Both MD17\cite{MD17} and revMD17\cite{revMD17} share the same configurations, but revMD17 improves upon MD17 by removing all the numerical noise present in the original data.
We utilized the revMD17 dataset as the ground truth but replaced the labels of 10\% of the configurations with values from the noisy MD17 dataset.
This introduces a source of systematic error distinct from random Gaussian noise and reminiscent of typical errors in electronic structure datasets.
We then trained MACE models on this mixed dataset, both with and without the proposed bootstrapping mechanism.

Fig.~\ref{fig:weights} shows the progress of the training run with bootstrapping, to illustrate the training dynamics of our proposed method.
In particular, we compare the loss distribution for early, middle and late epochs.
In early epochs it is difficult to distinguish between noisy and clean data.
As the model performance improves the noisy data clearly separates as outliers.
The weights of each configuration as a function of error included in panel b show how noisy data is effectively down-weighed over the course of the training process.

We next demonstrate in Fig.~\ref{fig:overfit} that bootstrapping effectively prevents the model from fitting the noisy labels.
In the standard training regime (without bootstrapping), the training error on the noisy subset decreases continuously.
However, the error with respect to the hidden ground truth (revMD17) eventually increases after reaching a minimum, confirming that the model is overfitting to the incorrect MD17 labels.

In contrast, the bootstrapped model successfully identifies these outliers.
As shown in Fig~\ref{fig:weights}~b), the loss weights for the noisy configurations rapidly drop toward zero by the 10th epoch.
Consequently, the training error on this subset plateaus at approximately 300 meV/Å, which corresponds to the inherent discrepancy between the MD17 and revMD17 evaluations.
Crucially, while the model attenuates the noisy training labels (by setting $w_i \approx 0$), its predictions for these configurations converge accurately toward the unseen ground-truth forces.

Finally, preventing overfitting on the noisy outliers significantly improves performance on the clean data.
As training progresses, the standard model degrades on the validation set due to the noise-induced distortion of the potential energy surface.
Conversely, the bootstrapped model maintains high accuracy, achieving a final median force error of 27.0 meV/Å on the re-evaluated validation set, which is a more than three-fold improvement over the 94.0 meV/Å error observed in the standard model.

\begin{figure}[tb]
    \centering
    \includegraphics[width=0.995\linewidth]{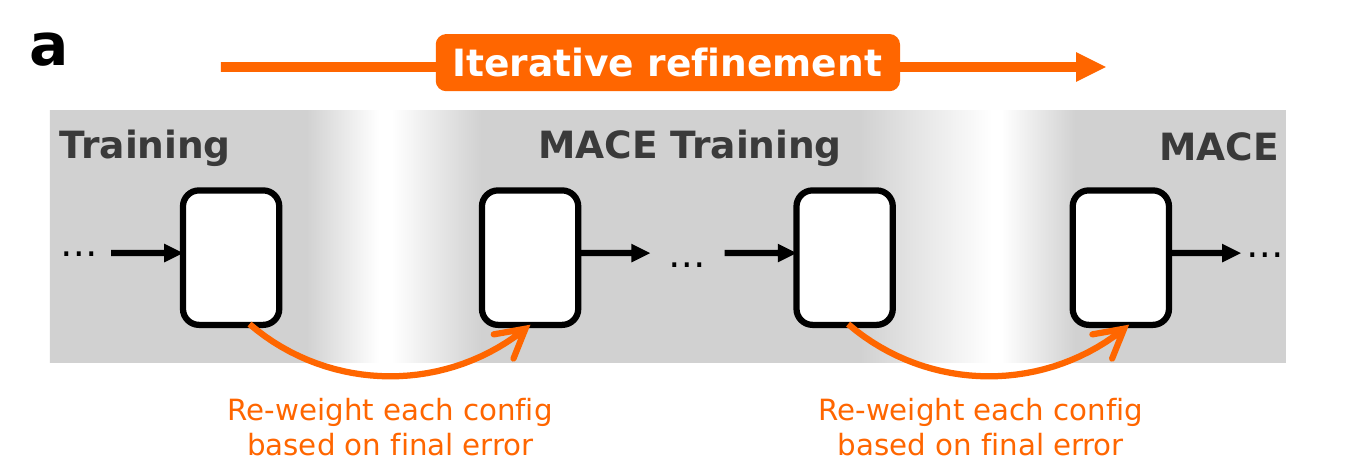}
    \includegraphics[width=0.945\linewidth]{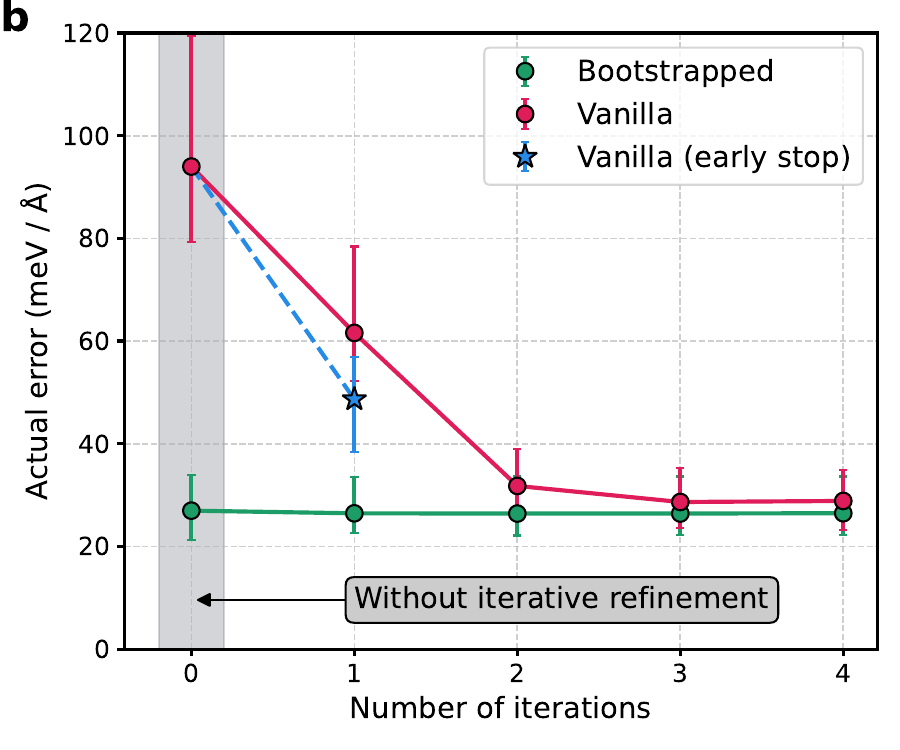}
    \caption{%
    \textbf{Noise-resilient training compared to iterative refinement.}
    (a) The schematics of iterative refinement , which involves re-evaluating the weights and then repeating the training for multiple cycles. The weight calculation is the same as bootstrapping without the need for an EMA. (b) The first model (step 0) is trained respectively with (green, `bootstrapped') and without (red, `vanilla') bootstrapping.  Distillation is done 4 times.  Force error with error bars (median and IQR) at the end of the 2000 epochs is plotted. Bootstrapping reaches the minimum force error ($\sim30\text{ meV/Å}$) without requiring refinement, while at least 2 refinement steps were required for a normal MACE model to reach this accuracy. A one-step refinement is also performed by stopping the initial training early at the 240th epoch (blue).      }
    \label{fig:distillation}
\end{figure}

\begin{figure*}[tb]
    \centering
    \includegraphics[width=\linewidth]{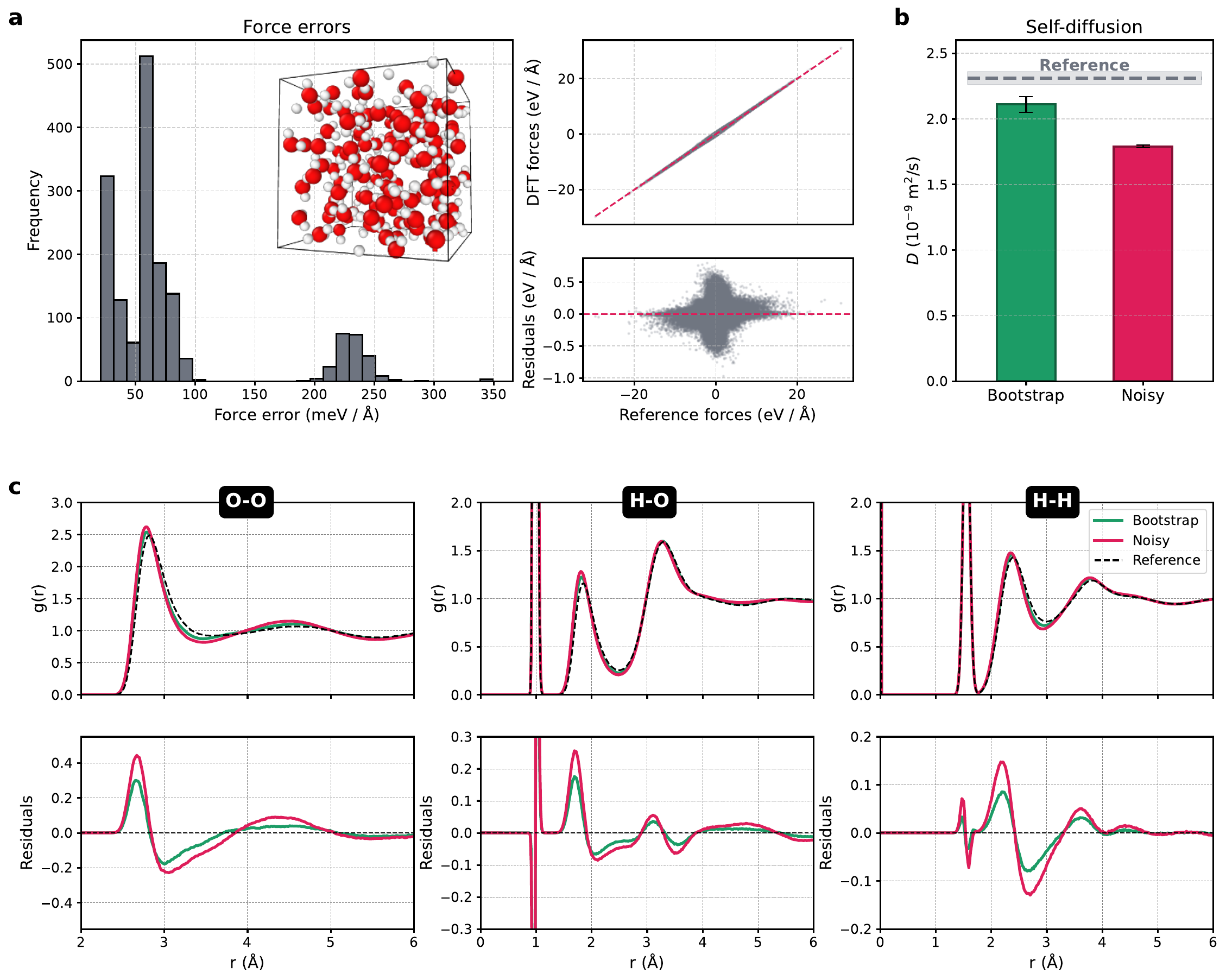}
    \caption{%
    \textbf{Effect on observables of bulk water simulations.}
    (a) (left) Histogram plot of RMSE forces produced by DFT in a certain loose convergence threshold and (right) the distribution of the forces and the residuals. (b) The self-diffusion coefficient of water at 298 K and 1 atm obtained in the classical MD simulation in NVT conditions, for a noisy dataset (red), and the noisy dataset trained with bootstrapping (green). Reference value computed from a clean dataset indicated as dotted line. (c) Radial distribution functions for (left) O-O, (middle) O-H, and (right) H-H interatomic distances in bulk liquid water at 298 K and 1 atm, taken from a classical MD simulation in NVT conditions.  Black dotted line plots the reference value. Residuals shown at the bottom.}
    \label{fig:Observables}
\end{figure*}

\subsection{Bootstrapping outperforms iterative refinement}

To assess the computational efficiency of our on-the-fly approach, we benchmarked it against a standard iterative refinement strategy~\cite{MACEMP0, MACEOFF}.
As illustrated in Fig.~\ref{fig:distillation}~a), this baseline method relies on the assumption that a trained model will exhibit high prediction errors on noisy outliers due to smoothness constraints.
The procedure involves training an initial model on the entire dataset, calculating the prediction errors for all training configurations, and assigning static weights based on these errors via Eq.\ (\ref{eq:weight_calc}).
A subsequent model is then trained on this re-weighted dataset.
We repeated this full training and re-weighting cycle over four generations to track the convergence of the data cleaning process.

Fig.~\ref{fig:distillation}~b) compares the validation errors of the iterative generations against our single-shot bootstrapped model.
While both methods eventually converge to a similar error of approximately 25--30 meV/Å, our approach achieves this accuracy in a single training run.
In contrast, the iterative baseline requires multiple computationally expensive retraining cycles to reach comparable performance.
For completeness, we also refine the bootstrapped model over multiple iterations, following the same procedure of removing high outlier data points after each run.
As visible in Fig.~\ref{fig:distillation}~b), the performance of the model doesn't improve nor degrade.
This shows that once the dataset has been filtered using the bootstrapping method by setting $w_i \approx 0$ to noisy samples, repetitive fittings do not result in the bootstrapping erroneously down-weighing clean samples. 

Furthermore, the iterative approach is highly sensitive to the stopping criteria of the previous model.
If the previous model is trained too long without noise filtering, it overfits the outliers, resulting in incorrect weights for the next model.
The boostrapping approach does not suffer from the same limitations.
By continually updating outlier probabilities during the training process, the dynamic bootstrapping method eliminates the need for such manual tuning and prevents the initial overfitting of noisy data.
Based on these findings, we test the iterative refinement baseline, but when stopping the training early.
As visible in Fig.~\ref{fig:distillation}~b) (blue line) this improves the performance slightly. 
Nevertheless, it does not reach the bootstrapped accuracy and incurs almost twice the computational cost.
Furthermore it requires a manual selection of "early stopping criteria".
We show in Appendix~\ref{subsec:app-threshold-sensitivity}, that the bootstrapping approach is not very sensitive to the threshold, and hence requires minimal care when setting the hyper-parameters. 

\subsection{Bootstrapping improves physical observables}
\label{sec:phys-ob}

Benchmarking MLIPs solely on validation set errors is often insufficient, as these metrics do not guarantee the stability or accuracy of derived physical observables during molecular dynamics (MD) simulations~\cite{Tuckerman_2023, Fu2023, Stocker_2022}.
To assess the practical utility of noise-resilient training, we evaluated the recovery of macroscopic properties of liquid water, for a system of 126 water molecules, relative to a clean dataset.
This allows the approach to be tested in a condensed phase system rather than on gas phase cluster as in the previous section.

\begin{figure*}[t]
\includegraphics[width=0.66\linewidth]{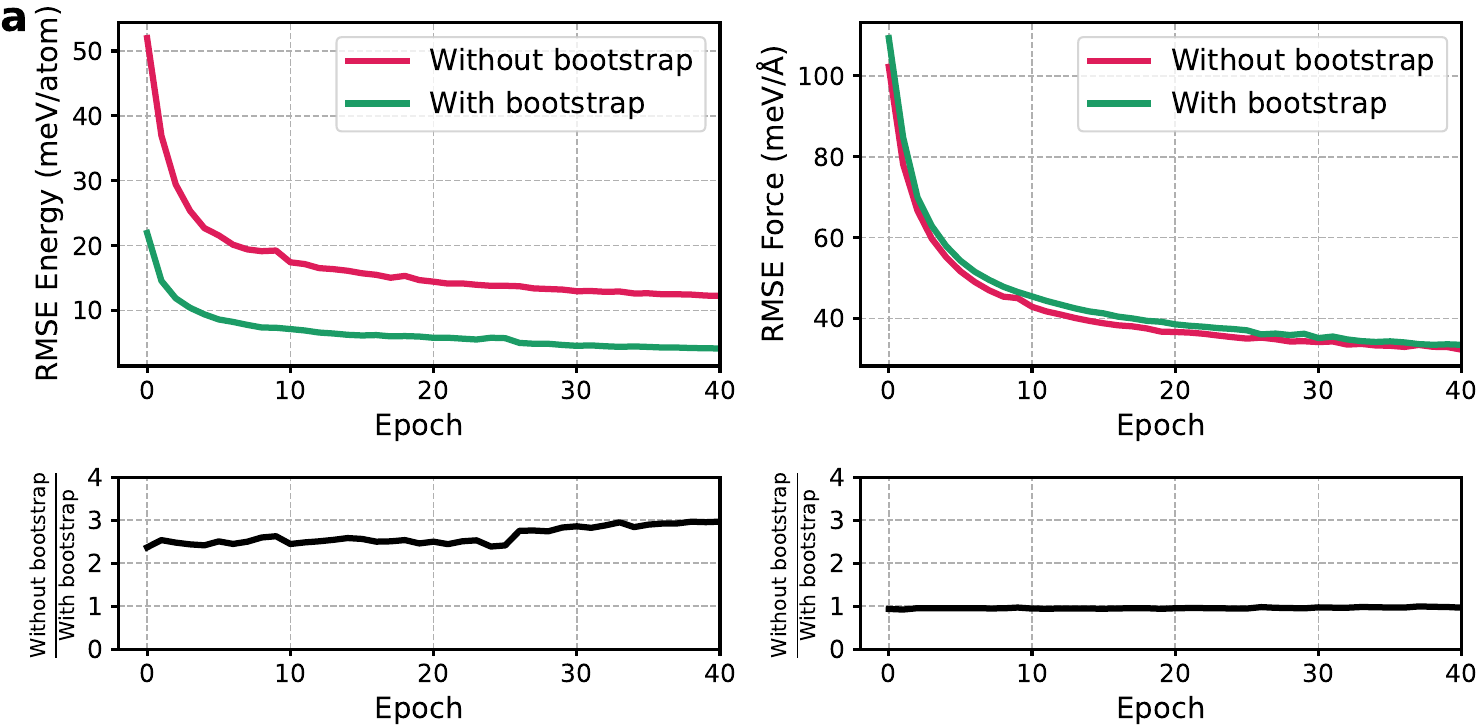}
\includegraphics[width=0.33\linewidth]{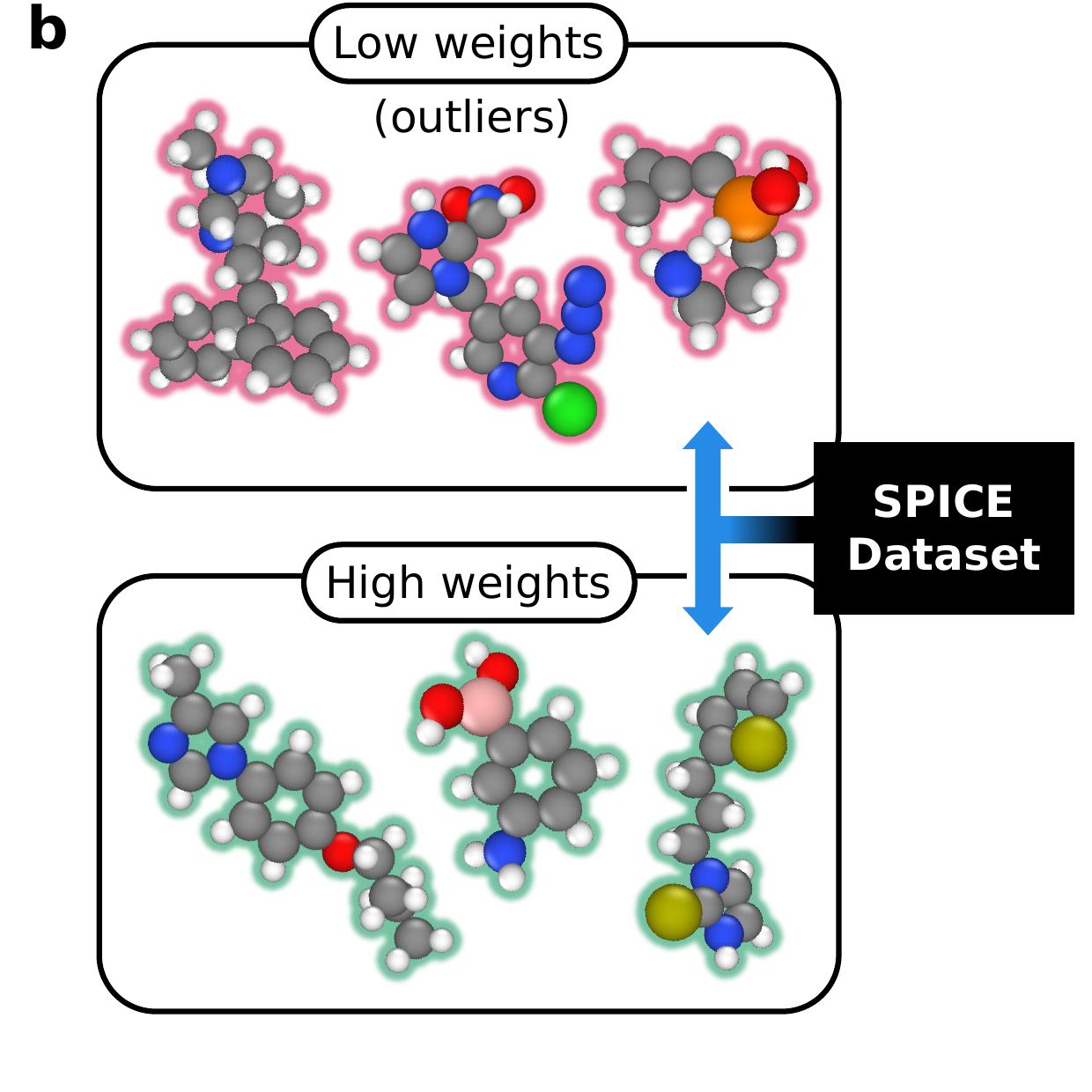}
    \caption{%
    \textbf{Noise-resilient training applied to the SPICEv2~\cite{SPICE2} dataset.}
    (a) Top panels: The evolution of the RMSE energy and forces training with boostrapping (green) and without bootstrapping (red) in the training lasting 40 epochs.  Each quantity is averaged over an epoch, so the value . Bottom panels: The ratio of error without bootstrapping relative to with bootstrapping. The ratio increases steadily up to 3 in energy, and maintains a value close to 1 in forces.  (b) Examples of molecules from the SPICEv2~\cite{SPICE1, SPICE2} dataset with different per-config weights.  Here ones with particularly low weights (i.e.\ classified as outliers) and high weights are shown, showcasing the ability for on-the-fly noise detection to filter configurations.     }
    \label{fig:spice}
\end{figure*}

To emulate a realistic scenario where high-throughput DFT calculations fail to fully converge, we generated a ``noisy'' dataset using loose SCF convergence thresholds. 
We contrast this with a ``clean'' dataset generated using tight convergence settings.
A comparison of the two reveals a multi-modal error distribution in the forces (Fig.~~\ref{fig:Observables}~a)), where approximately $12\%$ of the loosely converged configurations exhibit large force errors ($\gtrsim 200 \text{ meV/Å}$), classifying them as outliers.
For our subsequent assessment, we trained three models: a \textit{reference} model (trained on clean data), a \textit{standard} model (trained on noisy data), and a \textit{bootstrapped} model (trained on noisy data with our outlier detection scheme).

We first examined the dynamical properties of the system by computing the self-diffusion coefficient, $D$, from the mean squared displacement (Fig.~\ref{fig:Observables}~b))~\cite{ernst2013measuring}.
The \textit{reference} model yields a value of $(2.31 \pm 0.05)\times10^{-9} \text{ m}^2/\text{s}$.
The \textit{standard} model significantly underestimates diffusivity, measuring $(1.79 \pm 0.01)\times10^{-9} \text{ m}^2/\text{s}$, exhibiting a deviation of $0.52\times10^{-9} \text{ m}^2/\text{s}$.
In contrast, the \textit{bootstrapped} model recovers a value of $(2.11 \pm 0.06) \times10^{-9} \text{ m}^2/\text{s}$, deviating from the reference by only $0.20\times10^{-9} \text{ m}^2/\text{s}$.
This demonstrates that force errors have a significant impact on dynamical observables and that the bootstrapped model successfully filters the unphysical forces improving diffusion.

This improvement extends to structural properties, as quantified by the radial distribution functions (RDFs) for O-O, O-H, and H-H pairs, shown in Fig.~\ref{fig:Observables}~c).
The \textit{standard} model predicts a slightly over-structured liquid, a known artifact of overfitting to erroneous forces~\cite{Chen_2017}.
The \textit{bootstrapped} model corrects this behavior, reducing the residuals with respect to the reference RDF, particularly near the coordination peaks, while differences remain.
These results confirm that noise-resilient training allows for the recovery of accurate dynamical and correction of structural properties from unconverged training data, preventing the propagation of systematic electronic structure errors into macroscopic observables.

\subsection{Bootstrapping applies to foundation models}
Finally, we demonstrate the utility of on-the-fly outlier detection in the regime of foundation models.
While these models exhibit broad generalizability across chemical space, their accuracy is often compromised by the inclusion of outliers and unphysical structures inherent to large-scale datasets~\cite{Focassio_2025,Kuryla_2025}.
As the diversity of the training data increases, the probability of encountering artifacts from failed or unconverged \textit{ab initio} calculations rises.
Eliminating these data points is critical to prevent model degradation.

We applied our dynamic bootstrapping approach to the training of a MACE foundation model on the SPICE dataset~\cite{SPICE1, SPICE2}.
This dataset contains over $2.0$ million configurations of drug-like molecules and proteins spanning 17 chemical elements.
Previous assessments have noted quality issues within SPICE, including configurations with significant non-zero net forces and large errors associated with heavier halogens~\cite{Kuryla_2025}.

In Fig.~\ref{fig:spice}~a) we compare the performance of the bootstrapped model against a standard baseline over $40$ training epochs.
In the absence of a re-computed ``clean'' reference for this scale of data, we rely on validation RMSE to assess convergence stability.
We take select configurations from Kuryla et al.\ cleaned subset as the validation set~\cite{Kuryla_2025}. 
The bootstrapped model exhibits a significant improvement in energy prediction accuracy, achieving a validation RMSE $2.5$ times lower than the baseline in the first epoch.
This performance gap widens to a factor of $3.0$ by the 40th epoch.
The immediate divergence of the loss curves highlights the efficiency of the online filtering, which identifies outliers within the first pass of the dataset.
While the force RMSE remains comparable between the two methods (ratio $\approx 0.98$), the substantial reduction in energy error suggests that the bootstrapping successfully isolated high-energy outliers that otherwise dominate the loss function.

To validate the physical basis of this filtering, we examined configurations assigned low confidence weights $w_i$ by the model in Fig.~\ref{fig:spice}~b).
The method correctly identified unphysical structures, such as atoms with overlapping van der Waals radii or sterically clashing conformations, particularly in phosphorus-containing compounds.
These high-energy artifacts distort the learning of the interatomic interactions.
By systematically down-weighting these samples, the model avoids overfitting to unphysical geometries with difficult to converge electronic structure.
We note that while distinguishing between complex, ``hard-to-learn'' chemistry and erroneous data is non-trivial \cite{Smart_Carneiro_2023}, the magnitude of errors associated with steric clashes typically far exceeds that of valid but complex chemical environments.

These results indicate that noise-resilient training effectively scales to the foundation model regime.
The approach robustly identifies and mitigates the impact of outliers in large, diverse datasets, significantly improving validation accuracy without the need for manual curation.

\section{Conclusion}

In this work, we introduce an automated, on-the-fly outlier detection scheme that enables the robust training of MLIPs on noisy datasets.
By leveraging the distinct training dynamics of noisy versus clean data, we identify outliers based on their slower convergence rates relative to the noiseless data.
This approach systematically down-weights suspected outliers without requiring prior chemical knowledge or computationally expensive iterative retraining cycles.

Our results demonstrate that this dynamic bootstrapping prevents the model from overfitting to noisy samples, thereby achieving lower error rates on the underlying clean data.
Notably, the model is able to recover ground-truth forces despite being trained on noisy labels.
Furthermore, evaluations on a water dataset containing poorly converged DFT calculations reveal that filtering outliers corrects the prediction of physical properties in molecular dynamics simulations.
We observe improvement in the calculated self-diffusion constant and marginal improvements of radial distribution functions compared to the baseline.
Finally, we demonstrate the scalability of noise-resilient training to the foundation-model regime, where the method successfully filters unphysical chemical structures and reduced energy prediction errors.

This automated noise-resilience training procedure provides a new, practical solution for training on imperfect datasets with almost zero overhead, offering a pathway toward robust ML-driven molecular discovery without the need for extensive manual curation.

\section*{Computational Details}
\label{sec:comp-det}

In this section, we provide additional information on the training datasets and the proposed method.
Furthermore, we list detailed settings for the molecular dynamics simulations underlying Sec.~\ref{sec:phys-ob}. 

\subsection{Details of the Datasets}
\textbf{revMD17}: The original MD17 contained trajectories of ten molecules \cite{MD17}, and was re-evaluated in revMD17 with PBE/def2-SVP level of theory \cite{revMD17} using tighter SCF convergence plus denser DFT integration grid.
1000 spaced samples of Aspirin were chosen, and 10\% of the forces were replaced by the values computed in MD17.
Among the molecules, Aspirin was chosen for its relative complexity\cite{MACE}. 
The discrepancy in the forces of samples from the MD17 dataset show discrepancy with that from revMD17 between $250$ to $350 \text{ meV/Å}$.

\textbf{H$_\mathbf{2}$O}: Density Functional Theory (DFT) calculations were performed with FHI-AIMs \cite{FHI-aims}, on a periodic box (of size $15.835$ Å) containing 126 water molecules with the revPBE functional\cite{revPBE1, revPBE2}.
In contrast to a tight threshold of $\rho=10^{-3} \, e/a_0^3$ (charge density) and $F=10^{-5}\text{ eV/Å}$ (force) used to generate a reference dataset, we have used $\rho = 10^{3} \, e/a_0^3$ and $F=0.24 \text{ eV/Å}$ to emulate a poorly converged DFT setup.
In total 1616 configurations were generated, and errors are shown in Fig.~\ref{fig:Observables}~a).

\textbf{SPICE}:  The version 2.0 of the dataset was used, computed in the Psi4 software with the $\omega$B97M-D3(BJ) DFT functional and def2-TZVPPD basis set \cite{SPICE1, SPICE2}.
As a noise free validation set we use the manually filtered set from Ref.~\citenum{Kuryla_2025}.

\subsection{Training MACE}
To showcase the developed approach, we have implemented it for the MACE architecture, which utilises many-body messages and equivariant features\cite{MACE, MACE_design}, showing high data efficiency and robust performance~\cite{Kovacs_2023}. 

For the three datasets considered, we used different hyperparameters in MACE to reflect the complexity of the training data.
The MACE models trained on the revMD17 dataset and the water dataset comprise of 2 message passing layers with 128 channels and a 4 Å cutoff, as well as $L_\text{max}=1$ (equivariant messages).
As there is a constant offset in total energy around $\approx 950 \text{ meV}$ per atom per configuration overall in MD17 compared to revMD17, we trained only on the forces of the dataset. 
However, we train on both energies and forces in the water dataset, and stage two training was activated after 1500 epochs~\cite{Kovacs_2023}.
For each dataset, 15\% was held back as a validation set, and the training was run for 2000 epochs. 

For the training runs on the SPICE 2.0 dataset, 512 channels and a 6 Å cutoff were used.
4 NVIDIA H200 GPUs are employed in parallel to train for 24 hours.
In addition to the workflow given in Fig.~\ref{fig:schematic}~b), the statistics of the training process ($\mu$ and $\sigma^2$) are collected at the end of each epoch from all GPUs to take the average and redistribute, as each GPU will process different batches in parallel.
Given the simplicity of the outlier detection process, it will not add significant computational cost to the training of foundation models, which require sizeable effort to train to begin with.

\subsection{Molecular Dynamics Simulations}
With the resulting water MACE model, MD simulations were performed to obtain structural properties and diffusion coefficients. 
All MD simulations were performed using the Large-scale Atomic/Molecular Massively Parallel Simulator (LAMMPS) code \cite{Lammps}, coupled with the \texttt{Symmetrix} \cite{symmetrix} library, using the \texttt{symmetrix/mace} pair style.
All simulation boxes contained 126 water molecules and were performed at 298 K enforced by the CSVR thermostat with a temperature relaxation time of 0.1 ps~\cite{bussi2007canonical}.
Simulations in the canonical (NVT) ensemble used a 0.5 fs timestep, for a total of a 0.5 ns simulation. 
The diffusion coefficient was obtained from the mean squared displacement, fitting on a window between the first 0.01 ns to 0.1 ns. 

\begin{acknowledgments}

The authors thank Shoaib Ahmed Siddiqui for the initial discussions that inspired this project, as well as Rokas Elijošius, Ilyes Batatia, and Domantas Kuryla for their insights regarding noise filtering in MLIP training data.

T.C.W.L. kindly acknowledges support from the Cambridge Undergraduate Research Opportunities Programme and from Corpus Christi College, Cambridge for a Travel Grant.
N.O.N acknowledges financial support from the Gates Cambridge Trust.
C.S. acknowledges financial support from the Royal Society, grant number RGS/R2/242614, the UKRI Critical Mass grant, project reference EP/V062654/1, and the Isaac Newton Trust, award number G122390.
L.L.S. would like to acknowledge support from Wolfson College, Cambridge through a Junior Research Fellowship.

We are grateful for computational support and resources from the UK national high-performance computing service, Advanced Research Computing High End Resource (ARCHER2) through the APP59749: ML4HetCat project.
This work was also performed using resources provided by the Cambridge Service for Data Driven Discovery (CSD3) operated by the University of Cambridge Research Computing Service (www.csd3.cam.ac.uk), provided by Dell EMC and Intel using Tier-2 funding from the Engineering and Physical Sciences Research Council (capital grant EP/T022159/1), and DiRAC funding from the Science and Technology Facilities Council (www.dirac.ac.uk).
Access to CSD3 was obtained through a University of Cambridge EPSRC Core Equipment Award EP/X034712/1
\end{acknowledgments}

\section*{Data and Code Availability}

All datasets used for training and benchmarking are taken from the literature and publicly available.
Precise train, validation, test splits will be made available upon publication via Zenodo.
Noise-resilient training has been implemented as part of the MACE framework.
The code will be published as a pull-request to the main MACE code-base \url{www.github.com/ACEsuit/mace} upon publication.

\section*{Competing Interests}
The authors declare no competing interests.

\section*{References}

\putbib

\end{bibunit}

\appendix

\newpage

\section*{APPENDIX}

\section{Smooth weight estimation: EMA of \texorpdfstring{$\boldsymbol{\sigma}$}{sigma} and \texorpdfstring{$\boldsymbol{\mu}$}{mu} During Training Run}
\label{subsec:app-weight-estimation}

\begin{figure}[tbh]
    \centering
    \includegraphics[width=0.8\linewidth]{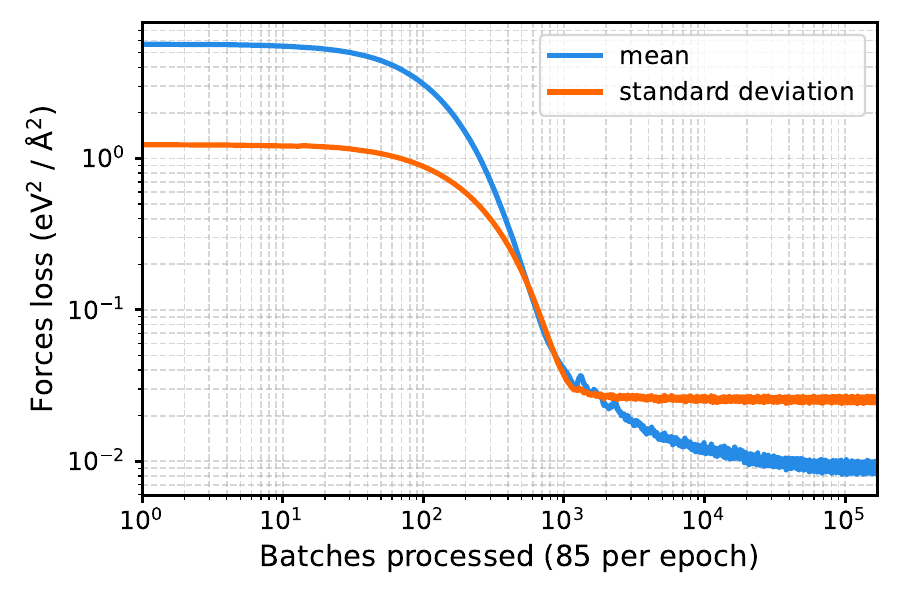}
    \caption{%
    \textbf{Smoothened average of the training statistics.}
    The exponential moving average of the mean and standard deviation of the forces loss as a function of the number of batches processed, trained on the revMD17 dataset, where in each epoch 85 batches are processed.}
    \label{fig:EMA}
\end{figure}

An exponential moving average is applied to the mean and standard deviation of the loss $\mathcal{L}_i$, as detailed in Eq. (\ref{eq:EMA1}) and (\ref{eq:EMA2}).  
We show how this leads to a smooth trend of these two quantities during training.  Figure~\ref{fig:EMA} shows the tracked statistics during the training of the revMD17 dataset.

\newpage

\section{Sensitivity Analysis of Bootstrapping Threshold \texorpdfstring{$\boldsymbol{z}_{\boldsymbol{t}}$}{z\_t}}
\label{subsec:app-threshold-sensitivity}

\begin{figure}[tbh]
    \centering
    \includegraphics[width=0.8\linewidth]{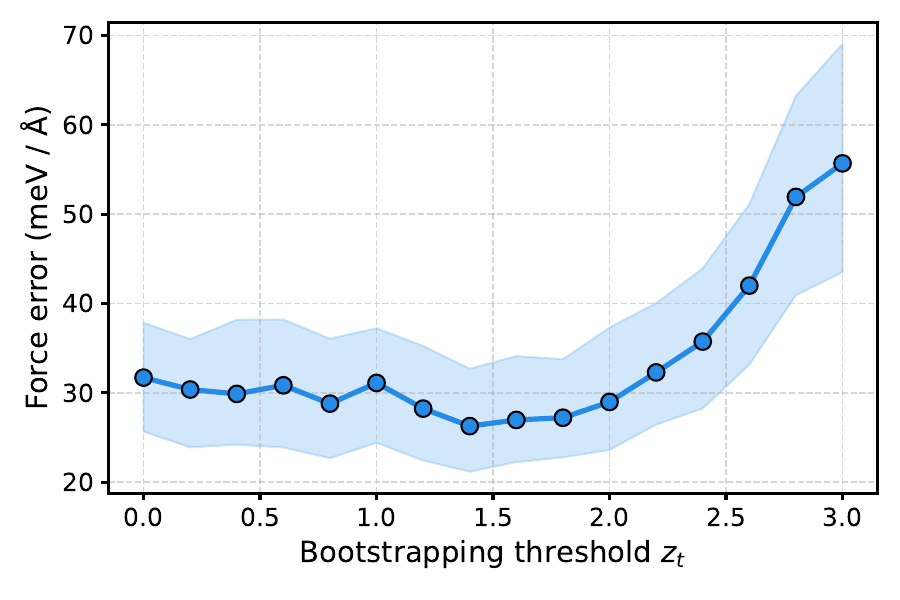}
    \caption{%
    \textbf{Sensitivity analysis of the bootstrapping threshold $\boldsymbol{z}_{\boldsymbol{t}}$.}
    The median of the final force error with respect to the re-evaluated revMD17 for bootstrapping thresholds in the range $[0.0, 3.0]$.  The shaded region represents the IQR. }
    \label{fig:Sensitivity}
\end{figure}

We trained models based on the same revMD17 dataset with on-the-fly outlier detection using a range of bootstrapping thresholds $z_t$ (see Eq. (\ref{eq:weight_calc}) for the definition) between $0.0$ and $3.0$.  The final force error (after 2000 epochs) with respect to the re-evaluated revMD17 set are shown in Figure~\ref{fig:Sensitivity}.  The graph shows that the effectiveness of bootstrapping is insensitive of the threshold for thresholds $z_t \lesssim 2.2$.  (Given that outliers comprise $10\%$ of the whole dataset, we would expect $z_t \approx 1.28$ to be most representative.)  This allows flexibility in $z_t$ without full knowledge of how noisy the dataset is.

\newpage

\end{document}